\documentclass[Royal,sageh,times]{sagej}

\fontsize{9}{11}\selectfont 
\usepackage{anysize}
\marginsize{0.8in}{0.8in}{0.8in}{0.8in}

\usepackage{moreverb,url}

\usepackage[colorlinks,bookmarksopen,bookmarksnumbered,citecolor=red,urlcolor=red]{hyperref}
\usepackage{algorithm}%
\usepackage{listings}%
\usepackage{longtable}
\usepackage{xpatch}
\usepackage{algorithmic} 
\usepackage{natbib}
\setcitestyle{numbers,square}
\newcommand\BibTeX{{\rmfamily B\kern-.05em \textsc{i\kern-.025em b}\kern-.08em
T\kern-.1667em\lower.7ex\hbox{E}\kern-.125emX}}

\def\volumeyear{2016}

\begin{document}

% \runninghead{Smith and Wittkopf}

\title{MTS: A Deep Reinforcement Learning Portfolio Management Framework with Time-Awareness and Short-Selling}

\author{Fengchen Gu\affilnum{1}, Zhengyong Jiang\affilnum{1}, Ángel F. García-Fernández\affilnum{2}\affilnum{3}, Angelos Stefanidis\affilnum{1}, Jionglong Su\affilnum{1} and Huakang Li\affilnum{1}}

\affiliation{\affilnum{1}School of AI and Advanced Computing, XJTLU Entrepreneur College (Taicang), Xi'an Jiaotong-Liverpool University, China\\
\affilnum{2}Department of Electrical Engineering and Electronics, University of Liverpool, UK\\
\affilnum{3}ETSI de Telecomunicaci\'on, Universidad Polit\'ecnica de Madrid, 28040 Madrid, Spain}

\corrauth{Jionglong Su and Huakang Li}

\email{Jionglong.Su@xjtlu.edu.cn, Huakang.Li@xjtlu.edu.cn}

\begin{abstract}
Portfolio management remains a crucial challenge in finance, with traditional methods often falling short in complex and volatile market environments. While deep reinforcement approaches have shown promise, they still face limitations in dynamic risk management, exploitation of temporal markets, and incorporation of complex trading strategies such as short-selling. These limitations can lead to suboptimal portfolio performance, increased vulnerability to market volatility, and missed opportunities in capturing potential returns from diverse market conditions. This paper introduces a Deep Reinforcement Learning Portfolio Management Framework with Time-Awareness and Short-Selling (MTS), offering a robust and adaptive strategy for sustainable investment performance. This framework utilizes a novel encoder-attention mechanism to address the limitations by incorporating temporal market characteristics, a parallel strategy for automated short-selling based on market trends, and risk management through innovative Incremental Conditional Value at Risk, enhancing adaptability and performance. Experimental validation on five diverse datasets from 2019 to 2023 demonstrates MTS's superiority over traditional algorithms and advanced machine learning techniques. MTS consistently achieves higher cumulative returns, Sharpe, Omega, and Sortino ratios, underscoring its effectiveness in balancing risk and return while adapting to market dynamics. MTS demonstrates an average relative increase of 30.67$\%$ in cumulative returns and 29.33$\%$ in Sharpe ratio compared to the next best-performing strategies across various datasets.
\end{abstract}


\maketitle

\section{Introduction}\label{s1}

Portfolio management is a crucial component of financial investment, aiming to optimize the balance between risk and return \cite{Tarasi2011Balancing}. A well-constructed portfolio can mitigate risks and capitalize on market opportunities, thereby achieving sustainable growth. Traditional portfolio management relies on diversification and periodic rebalancing to manage risk and enhance returns \cite{Kohler2014Rethinking}. However, the dynamic nature of financial markets necessitates more sophisticated approaches that can adapt to changing conditions \cite{Hommes2001Financial}. Market volatility, economic shifts, and unforeseen global events can all impact investment outcomes, making traditional static strategies insufficient \cite{Derbali2020Global}.

To address complex decision-making problems, the Deep Q-Network was introduced, marking the inception and widespread recognition of Deep Reinforcement Learning (DRL) \cite{Mnih2015HumanlevelCT}. DRL has emerged as a promising approach in portfolio management due to its ability to learn and adapt to complex environments \cite{Gao2020}. It integrates reinforcement learning with deep learning techniques, empowering algorithms to dynamically optimize strategies through continuous learning from historical patterns and real-time market feedback, thereby demonstrating robust adaptability to the inherent volatility and complex dynamics of financial markets.

Leveraging these advancements, various frameworks are developed to enhance trading strategies. The Ensemble of Identical Independent Evaluators (EIIE) and its variants use ensemble learning and policy gradients for cryptocurrency portfolios \cite{Jiang2018}. The Investor-Imitator (IMIT) mimics investor behavior for knowledge extraction \cite{2018Investor}. The FinRL framework supports various algorithms, such as Deep Deterministic Policy Gradient (DDPG), Soft Actor-Critic (SAC), and Proximal Policy Optimization (PPO), for single and multi-stock trading \cite{Liu2021}. The Ensemble Strategy (ES) selectively applies these algorithms to different time intervals \cite{Yang2020}. TradeMaster is an open-source platform for reinforcement learning-based trading that features a multi-strategy integration framework \cite{sun2023trademaster}. SARL integrates price movement predictions to enhance trading decisions \cite{Ye2020ReinforcementLearningBP}. Additionally, our previous work introduces a framework utilizing the Memory Instance Gated Transformer (MIGT) for effective portfolio management \cite{gu2025migtmemoryinstancegated}.

Despite the advancements in DRL-based portfolio management, the design of these strategies lacks comprehensive risk management and adaptability, which can be identified as three limitations. First, due to the reward structures used in training these models, most DRL-based portfolio management strategies overly emphasize return maximization, often at the expense of risk control, which might prioritize immediate gains over long-term stability, leading to potential large portfolio losses \cite{Campbell2001Optimal}. Some frameworks attempt to address risk management issues, but their approaches often rely on ex-post risk evaluation, such as liquidating all investments upon reaching a predefined risk threshold \cite{Liu2021, Li2024Research}. This unbalanced approach can result in algorithms that pursue extreme returns without adequately managing the accompanying risks, which leads to significant losses or heightened volatility. Second, existing strategies assume that market conditions are stationary and often do not account for temporal market characteristics, such as the weekend effect or the turn-of-the-month effect, which can significantly influence portfolio performance \cite{french1980stock, li2023volatility, Li2018Seasonality, kunkel2003turn, chiah2021tuesday}. Some studies employ the Fast Fourier Transform (FFT) to capture temporal or seasonal features, but FFT assumes that the input data is stationary, which does not align with the realities of the stock market \cite{Musbah2019Identifying}. Third, most portfolio management strategies assume that only long positions are allowed while short selling is prohibited, which does not reflect the realities of the stock market \cite{wu2021portfolio, 10076454}. This results in strategies failing to generate positive returns in declining or volatile markets. Despite short-selling being considered in some studies, thresholds and caps are mostly set manually or used only for risk hedging and without adequate risk controls, which can lead to significant potential for increased volatility and unforeseen losses\cite{asodekar2022deep, Chou2021Portfolio, Qiheng2023}.

The key contributions of our research are three-fold. First, we incorporate risk management into DRL algorithms using Incremental Conditional Value at Risk (ICVaR), addressing the limitation of overly emphasizing return maximization at the expense of risk control. These measures are integrated into the reward function to guide the algorithm towards better risk control, making the algorithms more adaptable to real-world scenarios. Second, we propose a new encoder and attention mechanism for DRL networks to incorporate temporal market characteristics and capture both long-term and short-term temporal patterns, enhancing their performance in varied market conditions. This innovation aims to develop robust and adaptive portfolio management strategies that better align with the complexities of real-world financial markets and effectively address the challenge of underutilizing temporal market characteristics. Third, we propose a parallel strategy for managing short-selling based on market trends, such as bull and bear markets. This strategy not only regulates the allowance of short-selling but also automates the control of risk aversion levels, enhancing the overall risk management framework. This comprehensive approach resolves the issues related to managing short-selling risks and limitations in existing strategies.

This paper is organized as follows: Section \ref{s2} provides the necessary definitions and assumptions for the portfolio management environment. In Section \ref{s3}, we present the proposed portfolio management framework, including the Markov decision process, risk controls, short-selling control framework, and time-aware embedding and attention. Experiments are conducted in Section \ref{s4} to evaluate the performance of our approach. Finally, we discuss the results and concludes the paper with future research directions in Section \ref{s5}.

\section{Definition}\label{s2}
\subsection{Transaction Process}\label{}
\begin{figure}
	\centering
	\includegraphics[width=14.5cm]{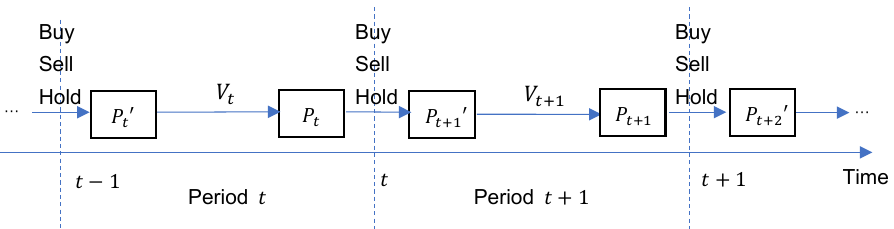}
	\caption{Interaction Process for Time Series. $V_t$ is the closing price vector in period $t$. $P_t$ is the portfolio value at the end of period $t$, and $P_{t+1}'$ is the portfolio value after it trades.}\label{fig1}
\end{figure}

In the DRL framework, the agent dynamically adjusts its capital allocation across stock classes at each discrete trading period $t\in \mathbb{N}^{+}$. Figure \ref{fig1} illustrates this portfolio management process. The model inputs a multidimensional tensor that includes historical time series of stock data and technical indicators at each period. The technical indicators contain Bollinger bands (BOLL) \cite{Murphy1999}, Commodity Channel Index (CCI) \cite{Altan2022}, Relative Strength Index (RSI) \cite{Altan2022}, True Range of Trading (TR) \cite{Chang2019}, Directional Movement Index (DMI) \cite{Seyma2020}, Moving Average Convergence Divergence (MACD) \cite{Hung2016}, and Money Flow Index (MFI) \cite{Singleton2014}. Each provides different market information independently of others, offering more comprehensive information for market analysis. We define each trading period $t$ to be one day. At the end of each period $t$, the agent executes trades on the portfolio based on $V_t$, which is a vector representing the closing prices of all stocks in that trading period and the total value of the portfolio denoted as $P_t$. This portfolio value includes the cumulative value of all securities in the portfolio and any cash reserves, providing a comprehensive valuation of the entire portfolio.

\subsection{Assumptions about the Experimental Environment}\label{}
For the constructed trading environment to be as realistic as possible, the following assumptions are made:
% Unnumbered list
\begin{itemize}
    \item Our simulated trades, grounded in historical data, are presumed not to influence stock prices. This premise is rooted in our trading volume being infinitesimal relative to the market's overall size, making any impact on stock prices negligible \cite{Soleymani2020, Jiang2018}.
    \item Given the considered daily trading frequency, each transaction price is set to the previous day's adjusted closing price, and the trading takes place in real time \cite{Liu2021}. Using the adjusted closing price accurately reflects the impact of dividends and stock splits on stock prices, ensuring data consistency and accuracy in investment decisions. Actions in period $t$ are classified as sell, buy, or hold. 
    \item To reflect various market expenses like trading and execution fees, we implement a universal transaction fee rate $c$ of 0.1$\%$ on each trade's value for both buying and selling operations \cite{Liu2021}. This standardization ensures compatibility across diverse trading contexts.
\end{itemize}

\section{Portfolio Management Framework}\label{s3}
\subsection{The Markov decision process of portfolio management}\label{}
\begin{figure}
	\centering
	\includegraphics[width=8cm]{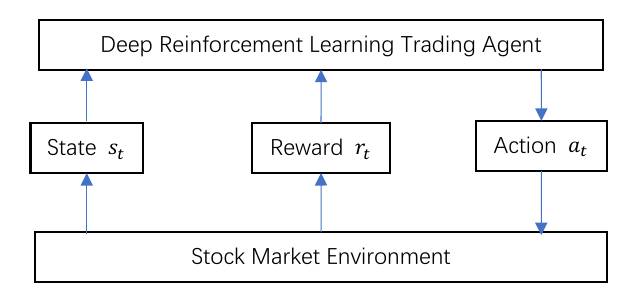}
	\caption{The Markov decision process of the DRL Environment, reflects the interaction of state, reward, action, DRL agents and the stock market environment.}\label{fig2t}
\end{figure}

The stock market is inherently stochastic, characterized by random and unpredictable movements in stock prices \cite{M.2022Recurrence}. To address the complexities of stock trading, we model the portfolio management task as a Markov Decision Process (MDP), as shown in Figure \ref{fig2t}. In this framework, the portfolio management problem is formulated as a decision-making process whose goal is to optimize the trading strategy.

In our MDP model, the state $s_t$ represents the current situation of the portfolio, including factors such as the current portfolio holdings, stock prices, and market conditions. At each period $t$, the agent, which is typically implemented using a Deep Reinforcement Learning (DRL) algorithm, chooses an action $a_t$ based on the current state $s_t$. Actions may include buying, selling, or holding various stocks in the portfolio. The agent then changes to a new state $s_{t+1}$ in response to the chosen action $a_t$, with the transition dynamics reflecting the inherent randomness of the stock market.

The training process of the DRL agent involves observing the changes in state $s_t$, executing actions $a_t$, and receiving rewards $r_t$, which are typically defined as the returns or profitability resulting from the actions taken. These rewards serve as a feedback signal, enabling the agent to evaluate the effectiveness of its actions. By repeatedly interacting with the environment and adjusting its strategy based on received rewards, the DRL agent learns to optimize its trading strategy.

The iterative nature of this learning process allows the DRL agent to refine its strategy over time. Through this reward-driven feedback loop, the agent can discover complex patterns and relationships in market data that are not immediately apparent. This process involves exploration of different actions and exploitation of known strategies to maximize cumulative rewards over time. The ultimate objective is to develop a trading strategy that generates the highest possible returns while managing risks and adapting to changing market conditions.

\begin{figure}
	\centering
	\includegraphics[width=16.5cm]{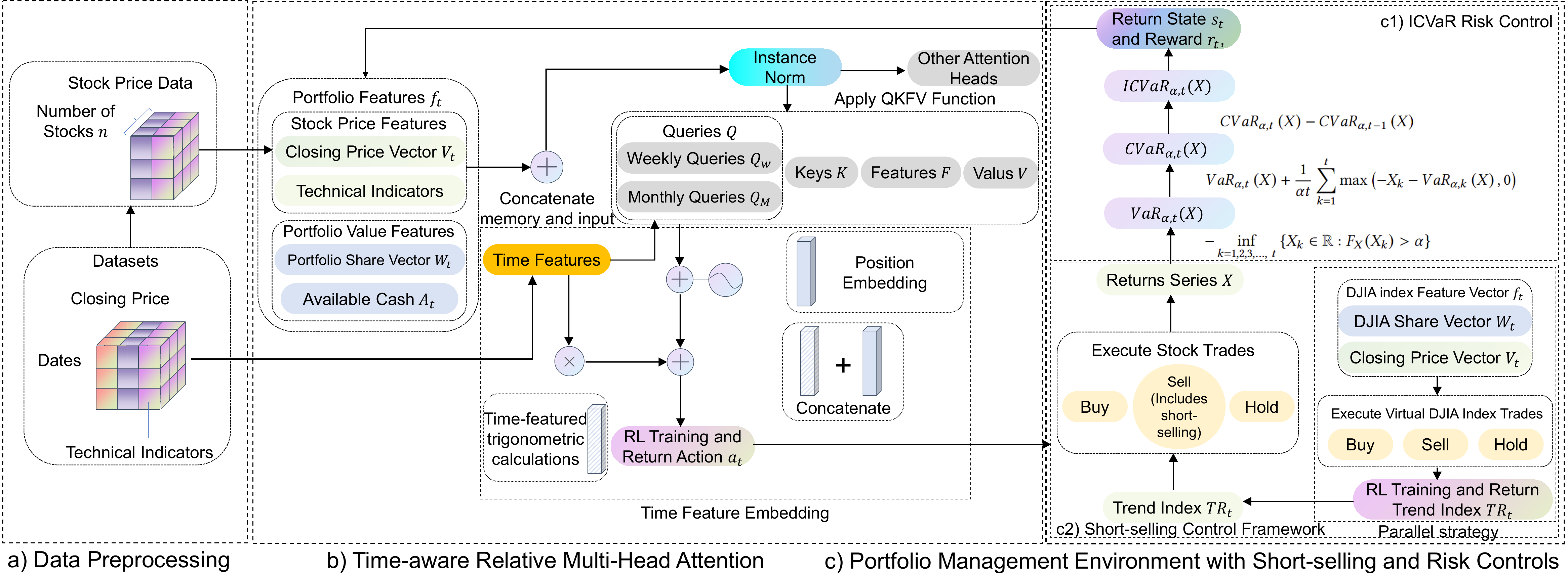}
	\caption{The proposed MTS framework for portfolio management. The first part is data input and preprocessing, which involves separating the time features and feeding them along with the stock data into the neural network. The second part involves the training of neural networks and reinforcement learning, where the core of the neural network is Time-Aware Embedding and Attention. It outputs actions $a_t$ for portfolio management and also receives outputs from the environment. The third part is the portfolio management environment, which consists of two main components. The first component is risk control that incorporates ICVaR. The second component is a stock trading mechanism that allows short selling and utilizes parallel strategies.}\label{figf}
\end{figure}

\subsection{Portfolio Management Environment with Risk Controls}\label{}
We first establish the DRL environment, including arrays representing stock prices and technical indicators. The state space is then defined to include elements such as the account cash amount, market volatility index, and technical indicators, using a Box space (a continuous range with defined upper and lower bounds for each variable) to indicate continuous states. The action space is defined where actions represent adjusting the weight of each stock, which is also continuous. The cash and stock holdings are randomly initialized to reset the environment, and the total stocks are calculated. Stock trades are executed based on actions, rewards, and new states calculated and returned. Portfolio details such as cash, holdings, and market indices are converted into state vectors to obtain the state. In addition, constraints are added, such as the minimum number of shares, to limit the impact of market volatility and transaction costs for trades.

The action $a_t$ is sell, buy, or hold for each stock, which consists of the target number of shares of each stock. We define $P_t\in \mathbb{R}^+$ as the portfolio value at the end of period $t$. The variable $P_t$ contains the available cash $A_t$ and the stock values $E_t$. The cash available in the portfolio at this period is:
\begin{equation}
	A_t=A_{t-1}+S_t^TV_t\left( 1-c\right)-B_t^TV_t\left(1+c\right), \label{e2}
\end{equation}
where $S_t$ is the share of stock sold in period $t$, $B_t$ represents the proportion of stock bought in period $t$, and $V_t$ is the closing price vector in period $t$. The length of the vectors $V_t$, $S_t$ and $B_t$ are the number of stocks $n$ in the portfolio. The portfolio share vector $W_t$ represents the share of each stock in the portfolio in period $t$, which is also the definition of action $a_t$. The state matrix in period $t$ is $s_t=\left[A_{t},f_{t,1}, f_{t,2}, \ldots, f_{t,n}\right]$, where $n$ is the number of stocks in the portfolio. Each element of $s_t$, $f_{t,k}=[ W_{t,k},\ V_{t,k},$ $ BOLL_{t,k}, CCI_{t,k}, RSI_{t,k}, TR_{t,k},$ $ DMI_{t,k}, MACD_{t,k}, MFI_{t,k}]$, is the feature vector of $k^{th}$ stock in period $t$, $k=1,2,3,\ldots,\ n$. If short-selling is allowed, $S_{t,n}$, the number of shares sold may be larger than they were before the transaction $W_{t-1,n}$, and $W_{t,n}$ can be negative, where $n$ represents the $n^{th}$ stock in the portfolio. $SH_{t,n}$ is the number of shares of the $n^{th}$ stock that sells short in period $t$,
\begin{equation}
        SH_{t,n}=\max \left( 0,S_{t,n}-W_{t-1,n} \right). 
\end{equation}

 At each time period $t$, the portfolio share vector $W_t$ is from the previous time period $W_{t-1}$ adds the shares purchased $B_t$ and subtracts the shares sold $S_t$,
\begin{equation}
	a_t=W_t= W_{t-1}+\,\,B_t-\,\,S_t. \label{e3-}
\end{equation}

The value of stocks $E_t$ at period $t$ is the product of the transposition of the portfolio share vector $W_t$, where the share of each stock is non-negative, and the vector of all stocks' closing prices $V_t$:
\begin{equation}
	E_t=\,\,W_t^TV_t^{\,\,}=\,\,\left( W_{t-1}+\,\,B_t-\,\,S_t \right) ^TV_t^{{
	}}. \label{e3}
\end{equation}
The portfolio's value in period $t$, $P_t$, can be obtained by summing equations (\ref {e2}) and (\ref {e3}), i.e.,
\begin{equation}
	P_t=\ A_t+\ E_t=\ A_{t-1}+\ {W_{t-1}}^TV_t-c{\ \left(B_t+S_t\right)}^TV_t.\label{e4}
\end{equation}
From equation (\ref {e3}) and (\ref {e4}), we obtain the value of the stocks in the portfolio in the last period $t-1$, i.e.,
\begin{equation}
	P_{t-1}=\ A_{t-1}+\ E_{t-1}=\ A_{t-1}+\ {W_{t-1}}^TV_{t-1}. \label{e5}
\end{equation}
Subtracting equation (\ref {e5}) from equation (\ref {e4}), we obtain the change in the portfolio value in period $t$,
\begin{equation}
	\bigtriangleup P_t=P_t-P_{t-1}=W_{t-1}^T\left( V_t-V_{t-1} \right) -c\left( B_t+S_t \right) ^TV_t. \label{e6}
\end{equation}

To enhance our model's ability to navigate complex and dynamic market conditions, we make our risk judgments and reward functions more sensitive to incremental changes in tail risk. Tail Risk refers to the probability of extreme, low-frequency market events that reside in the tails of a return distribution. As an integral component of our risk assessment and reward function, we propose Incremental Conditional Value at Risk (ICVaR) to effectively model and manage this risk. This risk metric builds upon two foundational concepts in financial risk management: Conditional Value at Risk (CVaR) and Incremental Value at Risk (IVaR) \cite{rockafellar2000optimization, tasche2003shortcut, cooper2018incremental}. Their base formula Value at Risk (VaR) is

\begin{equation}
    {VaR}_{\alpha,t}(X) = -\inf_{k=1,2,3,\ldots,\ t} \left \{ X_{k} \in
        \mathbb{R}: F_{X}(X_{k})>\alpha \right \},  \label{ev}
\end{equation}
where $X$ represents the returns series of our portfolio, $X_k$ represents the return of the portfolio at the time $k$, and $\alpha$ denotes the significance level, i.e., a crucial parameter that determines the confidence interval for our risk estimations. For a given confidence level $\alpha$ and period $t$, VaR is the negative value of the minimum value of the cumulative distribution $X_k$ for which the probability exceeds $\alpha$. The last element $X_t$ of returns series $X$ can be calculated from equations (\ref {e5}) and (\ref {e6}), i.e., 
\begin{equation}
X_t=\frac{\bigtriangleup P_t}{P_{t-1}}=\frac{W_{t-1}^T\left( V_t-V_{t-1} \right) -c\left( B_t+S_t \right) ^TV_t}{\ A_{t-1}+\ {W_{t-1}}^TV_{t-1}}. 
\end{equation}

In Equation (\ref {ev}), ${VaR}_{\alpha,t}\left( X \right)$ quantifies the maximum potential loss that a portfolio may incur over a specified period $t$ at a given confidence level $\alpha$. VaR serves as a fundamental risk metric, providing a benchmark for evaluating potential losses under normal market conditions. 

Building on VaR, we introduce the Conditional Value at Risk \cite{rockafellar2000optimization}, i.e.,
\begin{equation}
    {CVaR}_{\alpha,t}\left( X \right) = {VaR}_{\alpha,t}\left( X \right) + \frac{1}{\alpha t}\sum_{k=1}^t{\max}\left( -X_k-{VaR}_{\alpha,k}\left( X \right) ,0 \right).
\end{equation}
CVaR provides a nuanced risk measure by calculating the expected loss in scenarios where the loss potentially exceeds VaR. It is computed by adding VaR to the average of losses that surpass VaR, scaled by the factors $\frac{1}{\alpha t}$ and summed over the time horizon $T$.

While CVaR provides a more comprehensive view of potential losses beyond VaR, it does not account for the evolving nature of risk over time. As market conditions can change rapidly, a static measure may not be sufficient for real-time risk management. Therefore, to enhance our model’s responsiveness to market fluctuations and provide a more accurate assessment of changing risk profiles, we define ICVaR:
\begin{equation}
\begin{split}
    {ICVaR}_{\alpha,t}\left( X \right) 
    &= {CVaR}_{\alpha,t}\left( X \right) - {CVaR}_{\alpha,t-1}\left( X \right)
    \\&={VaR}_{\alpha,t}\left( X \right) - {VaR}_{\alpha,t-1}\left( X \right) \\&+ \frac{1}{\alpha t}\sum_{k=1}^t{\max}\left( -X_k-{VaR}_{\alpha,k}\left( X \right) ,0 \right)-\frac{1}{\alpha (t-1)}\sum_{k=1}^{t-1}{\max}\left( -X_k-{VaR}_{\alpha,k}\left( X \right) ,0 \right).
\end{split}
\end{equation}
This metric captures the incremental change in risk by subtracting the CVaR of the previous period from the current period. This differential approach allows us to gauge how the portfolio's tail risk evolves, providing real-time and dynamic risk assessment that enables the portfolio management agent to respond flexibly to market fluctuations.

The environment moves to a new state based on the current state and action taken, with the reward function calculating the percentage change in total assets considering risk, as computed by the ICVaR averse utility function:

\begin{equation}
    r=\bigtriangleup P_t-\lambda {ICVaR}_{\alpha,t}\left( X \right)=W_{t-1}^{T}\left( V_t-V_{t-1} \right) -c\left( B_t+S_t \right) ^TV_t-\lambda {ICVaR}_{\alpha,t}\left( X \right),
\end{equation}
where $\lambda \in [0,\, \infty )$ is the risk aversion parameter. Total assets are defined as the amount of cash plus the holdings multiplied by the prices. Constraints include the minimum number of shares, transaction costs deducted, and the market halting trading due to anomalies. In this way, the state space, action space, reward calculation, episode termination, and constraints are defined to simulate the dynamics of portfolio management.

\subsection{Short-selling Control Framework}\label{}
The system architecture diagram of our framework is shown in Figure \ref{figf}. Central to the framework is the primary investment strategy (Algorithm \ref{a1}), determining and executing the final portfolio allocation and position deployment. Running in parallel is a sub-strategy which operates independently and is not directly involved in the final master portfolio management strategy but is used to generate analytical judgment and inform decision-making.

The overall system architecture of this DRL portfolio framework is as follows. The system begins by setting up the financial investment environment, configuring asset holdings, cash balances, and relevant parameters such as risk aversion and ICVaR parameters. A parallel sub-strategy operates to assess current market conditions and determine appropriate trading actions in real time. At each time step, the system retrieves market data and updates the internal state accordingly. The agent then uses a deep neural network to select an action based on the observed state, which may include long or short trading positions. After executing the chosen trade, the system updates the portfolio, recalculates the asset holdings and cash balance, and computes the total assets. Risk is managed by calculating ICVaR, which adjusts for risk aversion, ensuring that the agent takes risk-aware actions. Finally, the system calculates the reward for the agent, accounting for both profitability and risk, and then checks if the episode has ended before resetting for the next cycle.

We incorporate a parallel strategy (Algorithm \ref{a2}) into the main strategy to determine the market trend of the portfolio to determine whether to execute short-selling and its upper limit. It uses the same structure as the main strategy except that only the Dow Jones index and the cash account are used as portfolios. Furthermore, the objective of the reward function is to maximise returns, and there are no assumed fees because these represent a simplified model to gauge broader market trends, allowing for efficient tracking of overall market movement. So its reward function is simpler than the main strategy, i.e.,
\begin{equation}
    r=\bigtriangleup P_t=W_{t-1}^{T}\left( V_t-V_{t-1} \right).
\end{equation}
The output of Algorithm \ref{a2} is the trend index $TR_t\in \left[ 0,1 \right]$ in period $t$, calculated as the ratio of the position index value to the total assets, where the value of one and zero represents the maximum uptrend and the maximum downtrend respectively. The framework allows the main strategy to establish an appropriate number of short-selling positions based on the trend index. Such short-selling will be subject to an upper limit of $u$ as a risk management constraint. The limit $u$ is customizable; it can be set as a fixed value based on individual preferences or, as we have implemented, adjusted according to the virtual portfolio value of sub-strategies. For example, the initial $u$ is 1e4, when the sub-strategy reaches a desired return, e.g. a cumulative return of 5$\%$, short selling is suspended, and $u$ becomes zero. When the sub-strategy suffers a larger loss, such as a cumulative historical time-averaged return of less than -5$\%$, the value of $u$ is set to 1e5. So the position constraint for each short-selling transaction can be expressed as
\begin{equation}
	SH_{t,n}\le \frac{{u}\left( 1-{TR}_{{t}} \right)}{{V}_{{t,n}}}. \label{es}
\end{equation}
For risk control, the amount of buying will also be constrained by $TR_t$. So the total purchase per transaction is determined by available cash $A_t$, $TR_t$ and additional buying limit $v\in [0,1]$, i.e.,
\begin{equation}
	\sum_{t=1}^n{B_{t,n}}\le \frac{vA_tTR_t}{V_{t,n}}. \label{eb}
\end{equation}
Similarly, the additional buying limit $v$ can be adjusted by the virtual portfolio value of the sub-strategy. For example, if the cumulative return of the sub-strategy is less than -5$\%$, we set $v$ to 0.2. In other cases, it remains set to one.

\begin{algorithm}
\caption{Stock Trading Environment Step Function}
\begin{algorithmic}[1]
\STATE Run parallel DRL to obtain trend index $TR_t$ and strategy account status without short-selling and buying limits

\FOR{each stock to sell}
    \STATE Calculate the number of selling shares $\mathnormal{S}_t$: 
    \[
    \mathnormal{S}_t \gets \sum_{n=1}^N (\min(\max(\mathnormal{W}_{t-1,n}, 0) +  \frac{{u}\left( 1-{TR}_{{t}} \right)}{{V}_{{t-1,n}}}, -\mathnormal{C}_{t,n}))
    \]where $t$ represents the time period, $n$ represents the $n^{th}$ stock in the portfolio, $\mathnormal{W}_t$ is the number of shares, $\mathnormal{V}_t$ is the close price and $\mathnormal{C}_t$ is the buy volume action output by the reinforcement learning algorithm
    \STATE Update number of shares from the previous time period $t-1$ $\mathnormal{W}_t$ and cash $\mathnormal{A}_t$:
    \[
    \mathnormal{W}_{t,n} \gets \mathnormal{W}_{t-1,n} - \mathnormal{S}_t
    \]
    \[
    \mathnormal{A}_t \gets \mathnormal{A}_{t-1} + (\mathnormal{V}_{t-1,n} \times \mathnormal{S}_t \times (1 - c))
    \] \\where $c$ is the transaction cost percentage

\ENDFOR

\FOR{each stock to buy}
    \STATE Calculate the number of buying shares $\mathnormal{B}_t$: 
    \[
    \mathnormal{B}_t \gets \min\left( \frac{vA_tTR_t}{V_{t,n}} , \mathnormal{C}_{t,n}\right)
    \] 
    \STATE Update number of shares $\mathnormal{W}_t$ and cash $\mathnormal{A}_t$:
    \[
    \mathnormal{W}_{t,n} \gets \mathnormal{W}_{t-1,n} + \mathnormal{B}_t
    \]
    \[
    \mathnormal{A}_t \gets \mathnormal{A}_{t-1} - \mathnormal{V}_{t,n} \times \mathnormal{B}_t \times (1 + c)
    \] \\where $c$ is the transaction cost percentage

\ENDFOR

\STATE Update state and the total asset $\mathnormal{T}_{t,n}$ during period $t$ for stock $n$: 
\[
\mathnormal{T}_{t,n} \gets \mathnormal{A}_t + \sum_{n=1}^N (\mathnormal{W}_{t,n} \times \mathnormal{V}_{t,n})
\]

\STATE Calculate return $\mathnormal{E}_t$ and update $\mathnormal{R}_t$: 
\[
\mathnormal{E}_t \gets \frac{\mathnormal{T}_t}{\mathnormal{T}_{t-1}} - 1
\]
\STATE \textbf{Return} state, reward, done, and info dictionary
\end{algorithmic}
\label{a1}
\end{algorithm}

Finally, the master strategy uses all reference information to make the final asset investment and portfolio construction decisions to achieve the predefined investment objectives. In this way, the independent sub-strategy serves as a valuable reference and benchmark for comparison. The main portfolio strategy remains the primary focus. It makes the final investment deployment, short-selling judgment, and execution decisions, as well as attributing all position control and risk adjustments to the main strategy.

\begin{algorithm}
\caption{Parallel DRL Environment for trend index $TR_t$}
\begin{algorithmic}[1]

\FOR{DJIA index to simulated sell}
    \STATE Calculate $\mathnormal{S}_t$: 
    \[
    \mathnormal{S}_t \gets \min(\mathnormal{W}_{t-1} , -\mathnormal{C}_{t})
    \] where $\mathnormal{W}_t$ is the DJIA number of shares
    \STATE Update DJIA number of shares $\mathnormal{W}_t$ and cash $\mathnormal{A}_t$:
    \[
    \mathnormal{W}_{t} \gets \mathnormal{W}_{t-1} - \mathnormal{S}_t
    \]
    \[
    \mathnormal{A}_t \gets \mathnormal{A}_{t-1} + (\mathnormal{V}_{t} \times \mathnormal{S}_t )
    \] where $\mathnormal{V}_t$ is the close price

\ENDFOR

\FOR{DJIA index to simulated buy}
    \STATE Calculate the number of buying shares $\mathnormal{B}_t$ and update $\mathnormal{W}_t$ and $\mathnormal{A}_t$: 
    \[
    \mathnormal{B}_t \gets \min\left(\frac{\mathnormal{A}_t}{\mathnormal{V}_{t}} , \mathnormal{C}_{t}\right)
    \] 
    \STATE Update DJIA number of shares and cash:
    \[
    \mathnormal{W}_{t} \gets \mathnormal{W}_{t-1} + \mathnormal{B}_t
    \]
    \[
    \mathnormal{A}_t \gets \mathnormal{A}_{t-1} - \mathnormal{V}_{t} \times \mathnormal{B}_t 
    \] 

\ENDFOR

\STATE Update the total asset during period $t$ for index $n$ $\mathnormal{T}_{t}$: 
\[
\mathnormal{T}_{t} \gets \mathnormal{A}_t + \mathnormal{W}_{t} \times \mathnormal{V}_{t}
\]

\STATE Calculate return $\mathnormal{E}_t$:
\[
\mathnormal{E}_t \gets \frac{\mathnormal{T}_t}{\mathnormal{T}_{t-1}} - 1
\]
\STATE Calculate the return $\mathnormal{R}_t$:
\[
\mathnormal{R}_t \gets (\mathnormal{T}_t - \mathnormal{T}_{t-1}) \times \mathnormal{X}_t
\] where $\mathnormal{X}_t$ is the reward scaling
\STATE Calculate trend index $\mathnormal{TR}$: 
\[
\mathnormal{TR_t} \gets \frac{\mathnormal{W}_{t} \times \mathnormal{V}_{t}}{\mathnormal{T}_t}
\]
\STATE \textbf{Return} $\mathnormal{TR_t}$, state, reward, done, and info dictionary
\end{algorithmic}
\label{a2}
\end{algorithm}

\subsection{Time-Aware Embedding and Attention}\label{}

\begin{algorithm}
\caption{Time-aware Relative Multi-Head Attention}
\begin{algorithmic}[1]
\STATE Initialize parameters: number of heads $\mathnormal{H}$, head dimension $\mathnormal{D}$, input dimension $\mathnormal{I}_{dim}$, output dimension $\mathnormal{O}_{dim}$, activation function $f_{act}$
\STATE Initialize learnable parameters: $\mathnormal{U}, \mathnormal{W}, \mathnormal{Y} \in \mathbb{R}^{\mathnormal{H} \times \mathnormal{D}}$ \\
\STATE Define output projection: $L_{out}: \mathbb{R}^{\mathnormal{H} \times \mathnormal{D}} \to \mathbb{R}^{O_{dim}}, x \mapsto f_{act}(W_{out}x + b_{out})$
\\ where $W_{out} \in \mathbb{R}^{O_{dim} \times \mathnormal{H} \times \mathnormal{D}}, b_{out} \in \mathbb{R}^{O_{dim}}$
\STATE Define position projection: $L_{pos}: \mathbb{R}^{I_{dim}} \to \mathbb{R}^{\mathnormal{H} \times \mathnormal{D}}, x \mapsto W_{pos}x$, where $W_{pos} \in \mathbb{R}^{\mathnormal{H} \times \mathnormal{D} \times I_{dim}}$
\STATE Initialize learnable weights for combining scores: $w_c \in \mathbb{R}^3$
\STATE Input tensor $\mathnormal{X} \in \mathbb{R}^{\mathnormal{B} \times \mathnormal{T} \times \mathnormal{I}_{dim}}$ and memory $ \mathnormal{M} \in \mathbb{R}^{\mathnormal{B} \times \mathnormal{\tau} \times \mathnormal{I}_{dim}}$ 
\\ where $\mathnormal{B}$ is batch size and $\mathnormal{T}$ is time steps, $\mathnormal{\tau}$ is length of memory
\STATE Concatenate memory and input: $X \gets [M; X] \in \mathbb{R}^{B \times (T+\tau) \times I_{dim}}$
\STATE Apply Instance Normalization: $X \gets \frac{X - \mu(X)}{\sqrt{\sigma^2(X) + \epsilon}}$
\\ where $\mu(X), \sigma^2(X)$ are mean and variance along feature dimension
\STATE Project to queries, keys, time features, values: 
$[Q; K; F; V] \gets W_{qkfv}X + b_{qkfv}$
\\where $W_{qkfv} \in \mathbb{R}^{4\mathnormal{H} \times \mathnormal{D} \times I_{dim}}, b_{qkfv} \in \mathbb{R}^{4\mathnormal{H} \times \mathnormal{D}}$
\STATE Reshape and normalize: $Q, K, F, V \in \mathbb{R}^{B \times (T+\tau) \times H \times D}$\\
\begin{center}
$Q \gets \text{softmax}(Q, \text{dim}=2), K \gets \text{softmax}(K, \text{dim}=2)$
\end{center}
\begin{center}
$F \gets \text{softmax}(F, \text{dim}=2), V \gets \text{softmax}(V, \text{dim}=2)$
\end{center}
\STATE Define weekly and monthly masks:
\begin{center}
$
M_w[d] = \begin{cases}
1, & \text{if } d \bmod 5 = 0 \\
0, & \text{otherwise}
\end{cases}  \forall d \in {0, 1, ..., D-1} 
$
$
M_m[d] = \begin{cases}
1, & \text{if } d \bmod 21 = 0 \\
0, & \text{otherwise}
\end{cases}  \forall d \in {0, 1, ..., D-1}
$
\end{center}
where $M_w, M_m \in {0,1}^D$
\STATE Apply masks:
\[
Q_w \gets Q \odot M_w^{(B, T+\tau, H, 1)}, 
Q_m \gets Q \odot M_m^{(B, T+\tau, H, 1)}
\]
\\where $\odot$ is the symmetric product, $M_w^{(B, T+\tau, H, 1)}$ and $M_m^{(B, T+\tau, H, 1)}$ are $M_w$ and $M_m$ broadcast to shape $(B, T+\tau, H, D)$
\STATE Normalize masked queries:
\begin{center}
$
Q_w \gets \text{softmax}(Q_w, \text{dim}=2), 
Q_m \gets \text{softmax}(Q_m, \text{dim}=2)
$
\end{center}
\STATE Time feature encoding:
$ R_{time} \gets L_{pos}(E_{tf}(\tau + T)) \in \mathbb{R}^{(\tau+T) \times H \times D} $
\\where $E_{tf}$ is time feature embedding in Algorithm \ref{a4}
\STATE Compute attention scores:
\\\begin{center}
$
S_c \gets \langle Q + U, K \rangle \in \mathbb{R}^{B \times T \times (T+\tau) \times H}, \
S_w \gets \langle Q_w, F \rangle \in \mathbb{R}^{B \times T \times (T+\tau) \times H},$ \\
$S_m \gets \langle Q_m, F \rangle \in \mathbb{R}^{B \times T \times (T+\tau) \times H}, \
S_p \gets \langle Q + Y, R_{time} \rangle \in \mathbb{R}^{B \times T \times (T+\tau), \times H}
$
\end{center}
where $\langle \cdot, \cdot \rangle$ denotes the dot product along the last dimension
\STATE Combine scores:
\\\begin{center}
$
S \gets \frac{e^{w_c[0]}}{\sum_i e^{w_c[i]}} S_c + \frac{e^{w_c[1]}}{\sum_i e^{w_c[i]}} S_w + \frac{e^{w_c[2]}}{\sum_i e^{w_c[i]}} S_m + \text{RelShift}(S_p)
$
\end{center}
where $\text{RelShift}$ is the relative shift operation
\STATE Normalize scores: $S \gets \text{softmax}(S / \sqrt{D}, \text{dim}=2)$
\STATE Define causal mask: $M_{seq} \in {0,1}^{T \times (T+\tau)}$
\\ where $M_{seq}[i,j] = 1$ if $j \leq i + \tau$, else $0$
\STATE Apply causal mask: $S_{masked} \gets S \odot M_{seq} $
\STATE Compute attention weights: $W_{attn} \gets \text{softmax}(S_{masked}, \text{dim}=2)$
\STATE Compute weighted sum: $O \gets \sum_{j=1}^{T+\tau} W_{attn}[:,:,j,:] V[:, j, :, :]$
\STATE Reshape output: $O \gets O.\text{reshape}(B, T, \mathnormal{H} \times \mathnormal{D})$
\STATE \textbf{return} $L_{out}(O)$
\end{algorithmic}
\label{a3}
\end{algorithm}

The novel Time-aware Relative Multi-Head Attention algorithm (Figure \ref{figf}b and Algorithm \ref{a3}) introduces advancements in processing sequential data with complex temporal relationships. The Fast Fourier Transform (FFT) is commonly used for analyzing frequency components in time series data \cite{Fumi2013Fourier}. However, FFT has limitations in capturing non-stationary patterns and complex temporal dependencies, making it difficult to process real-world data that exhibit irregular and multi-scale temporal patterns \cite{Musbah2019Identifying}. The Time-aware Relative Multi-Head Attention algorithm can model these intricate temporal dependencies directly in the time domain, allowing for a more comprehensive understanding of the data’s temporal structure.

At the core of this algorithm lies an innovative time feature embedding mechanism, which captures long-term and short-term temporal patterns with weekly and monthly effects. The Time Feature Embedding (Algorithm \ref{a4} and part of Figure \ref{figf}b) starts by initializing parameters for output dimension $\mathnormal{O}_{dim}$ and sequence length $\mathnormal{S}_{length}$. It then generates a vector $\mathnormal{O}_{range}$ with a weekly effect and calculates inverse frequencies $\mathnormal{I}_{freq}$. Subsequently, the core of this embedding lies in its forward pass, where it generates position inputs $\mathnormal{P}_{input}$ and calculates sinusoidal inputs $\mathnormal{S}_{input}$. Using both sine and cosine functions, the method produces position embedding $\mathnormal{P}_{embeddings}$ to encode the positional information of the input sequence, allowing the model to capture the order and relative positions. Incorporating temporal effects sets this approach apart from relative positional encoding. It defines periods for weekly $\mathnormal{P}_{5}$ and monthly $\mathnormal{P}_{21}$ cycles, corresponding to the number of working days per week and month, respectively. These are used to calculate weekly sine $\mathnormal{W}_{sin}$, weekly cosine $\mathnormal{W}_{cos}$, monthly sine $\mathnormal{M}_{sin}$, and monthly cosine $\mathnormal{M}_{cos}$ components. The final embedding concatenates these components, creating a rich representation that captures multi-scale temporal patterns. This approach offers a significant advantage over relative positional encodings by explicitly modelling recurring patterns at different time scales, which is crucial for many real-world sequential data problems \cite{Weerakody2021A}.

\begin{algorithm}
\caption{Time Feature Embedding: Generating Time-Aware Position Embeddings}
\begin{algorithmic}[1]
\STATE Initialize parameters: $\mathnormal{O}_{dim} \in \mathbb{N}$, $\mathnormal{S}_{length} \in \mathbb{N}$ \\
where $\mathnormal{O}_{dim}$ is output dimension, $\mathnormal{S}_{length}$ is sequence length
\STATE Generate $\mathnormal{O}_{range}$ with weekly effect: $\mathnormal{O}_{range} \gets x_i, x_i = 5i$, \\
where $i \in \mathbb{N} \text{ and } x_i \in [0, 5, 10,..., {O}_{dim}]$
\STATE Calculate inverse frequency: $\mathnormal{I}_{freq} \gets 1 / (10000^{x_i / \mathnormal{O}_{dim}})$,  \\
where $\mathnormal{I}_{freq}$ is inverse frequency
\STATE \textbf{forward} {$\mathnormal{S}_{length}$}
\STATE Generate position input: $\mathnormal{P}_{input} \gets x_i, x_i = 5i, \text{ where } i \in \mathbb{N} \text{ and } x_i \in [0, 5, 10,..., \mathnormal{S}_{length}]$ \\
 where $\mathnormal{P}_{input}$ is position input
\STATE Calculate sinusoid input: $\mathnormal{S}_{input[i][j]} \gets p_{input[i]} \cdot I_{freq[j]}$ \\
where $\mathnormal{S}_{input}$ is sinusoid input
\STATE Generate position embeddings: $\mathnormal{P}_{embeddings} \gets [\sin(\mathnormal{S}_{input}), \cos(\mathnormal{S}_{input})]$ \\
where $\mathnormal{P}_{embeddings}$ is position embeddings.
\STATE Define periodic effects: $\mathnormal{P}_{5} \gets 2\pi / 5$ \\
where $\mathnormal{P}_{5}$ is period 5, the number of working days per week
\STATE $\mathnormal{P}_{21} \gets 2\pi / 21$ \\
where $\mathnormal{P}_{21}$ is period 21, the average number of monthly working days
\STATE Calculate weekly sine: $\mathnormal{W}_{sin} \gets \sin(\mathnormal{P}_{5} \cdot \mathnormal{P}_{input})$ 
\STATE Calculate weekly cosine: $\mathnormal{W}_{cos} \gets \cos(\mathnormal{P}_{5} \cdot \mathnormal{P}_{input})$ 
\STATE Calculate monthly sine: $\mathnormal{M}_{sin} \gets \sin(\mathnormal{P}_{21} \cdot \mathnormal{P}_{input})$ 
\STATE Calculate monthly cosine: $\mathnormal{M}_{cos} \gets \cos(\mathnormal{P}_{21} \cdot \mathnormal{P}_{input})$ 
\STATE Concatenate embeddings: $\mathnormal{P}_{embeddings} \gets [\mathnormal{P}_{embeddings}, \mathnormal{W}_{sin}, \mathnormal{W}_{cos}, \mathnormal{M}_{sin}, \mathnormal{M}_{cos}]$
\STATE \textbf{Return} $\mathnormal{P}_{embeddings}$
\STATE \textbf{end forward}
\end{algorithmic}
\label{a4}
\end{algorithm}

The attention mechanism in this algorithm builds upon the multi-head attention structure of traditional Transformers but with several key enhancements. The process begins with the projection of the input into queries $Q$, keys $K$, time features $F$, and values $V$. Unlike existing Transformers \cite{Tay2020Efficient, Mishev2020Evaluation}, this algorithm introduces time-aware query variants by applying weekly $M_w$ mask and monthly $M_m$ mask to the queries, resulting in $Q_w$ and $Q_m$. These masked queries allow the model to focus on specific temporal patterns. The algorithm calculates four types of attention scores: content-based scores $S_c$ using the query-key dot product, weekly time-aware scores $S_w$ using masked weekly queries and time features, monthly time-aware scores $S_m$ using masked monthly queries and time features, and positional scores $S_p$ using queries and relative time encodings. These scores are then combined using weights $w_c$ learned using the attention mechanism, allowing the model to balance different attention mechanisms adaptively.

The advantages of our attention mechanism are threefold. First, it enables the model to capture and utilize temporal patterns at multiple scales simultaneously, addressing a key limitation of existing Transformers in handling time series data \cite{Tay2020Efficient}. Second, using masked queries $Q_w$ and $Q_m$ allows for more focused attention on specific temporal patterns, enhancing the model's ability to capture recurring events or cycles in the data. Third, the adaptive combination of different attention scores through learnable weights $\mathnormal{U}, \mathnormal{W}$ and $\mathnormal{Y}$ allows the model to adjust its focus based on the specific characteristics of the input data and the task at hand. This allows the algorithm to understand and exploit temporal patterns at different time scales for complex sequence data tasks.

\section{Experiments}\label{s4}

\subsection{Comparison with other approaches}\label{}
To test the performance of the new framework in real markets, we carrie out experiments using historical data, using 30 stocks from the Dow Jones Industrial Average (DJIA) as the portfolio. We use data sets for the five most recent full calendar years from 2019 to 2023 as test sets refer to Table \ref{tbl1}, respectively. We assess each variant using four key metrics: Cumulative returns, Sharpe ratio, Omega ratio, and Sortino ratio \cite{Sharpe1994, Rollinger2015, Keating2002}. 

To evaluate the portfolio management framework, it is a direct approach to compare cumulative returns:
\begin{equation}
	Cumulative\ rate\ of\ return\ =\ \frac{P_T\ -P_0}{P_0}\,
\end{equation}
where $P_0$ is the initial investment, $P_T$ is the final value, and $T$ is the total number of periods. This metric shows the overall profit generated but does not account for risk.
The Sharpe ratio is widely used to measure risk-adjusted returns:
\begin{equation}
	Sharpe\ Ratio\ =\ \frac{R_p-R_f}{\sigma_p},
\end{equation}
where $R_p$ is the portfolio return, $R_f$ is the risk-free return (set to 3\% per year in our study), and $\sigma_p$ is the standard deviation of portfolio returns. A higher Sharpe ratio indicates better risk-adjusted performance.
To assess downside risk, we use the Sortino ratio:
\begin{equation}
	Sortino\ Ratio\ =\ \frac{R_p-r}{\sqrt{\frac{1}{T}\sum_{t=0}^{T}\left(R_{p_t}-r\right)^2}},
\end{equation}
where $r$ is the minimum acceptable return (3\% per year in our study), $T$ is the total number of periods, and $R_{p_t}$ is the return for period $t$. This ratio focuses on returns below the minimum acceptable level, with a higher value indicating better downside risk-adjusted returns.
For a comprehensive risk-return evaluation, we use the Omega ratio:
\begin{equation}
	Omega\,\,Ratio\,\,=\,\,\frac{\int_r^{\infty}{\left( 1-F\left( x \right) \right)}dx}{\int_{-\infty}^r{F}\left( x \right) dx},
\end{equation}
where $F$ is the cumulative distribution function of returns. This ratio compares potential gains to potential losses, with a higher value suggesting a more favorable risk-return profile.

To assess the performance of our strategy, we compare it with traditional statistical and DRL strategies. Our traditional statistical strategies are based on mean reversion, trend following, cost optimization, and machine learning principles. Mean reversion strategies are based on the assumption that stock prices will revert to their historical averages over time:

\begin{itemize}
    \item Confidence Weighted Mean Reversion (CWMR) \cite{Li2011} models the portfolio vector as a Gaussian distribution, updating it sequentially according to the mean reversion trading principle.
    \item Online Moving Average Reversion (OLMAR) \cite{Li2012} leverages multi-period moving average regression to inform its strategy.
    \item Passive Aggressive Mean Reversion (PAMR) \cite{Lii2012} utilizes the mean reversion relationship in financial markets and applies online passive-aggressive learning techniques from machine learning.
    \item Robust Median Reversion (RMR) \cite{Huang2016} uses the mean reversion properties of financial markets and implements robust L1-median estimation to address outliers in mean reversion.
    Weighted Moving Average Mean Reversion (WMAMR) \cite{Gao2013} predicts stock price trends by taking a weighted moving average of stock prices.
\end{itemize} 

Cost optimization strategies focus on reducing transaction costs to improve overall returns. Transaction Costs Optimization (TCO) \cite{Li2018} combines L1 parametrization of the difference between consecutive allocations with the principle of maximizing expected logarithmic returns, accommodating non-zero transaction costs. Trend-following strategies identify and capitalize on the momentum of stock prices. BEST \cite{Jiang2017} selects the stock with the best performance from the previous day. Machine learning strategies utilize advanced algorithms to identify patterns and make predictions based on data. Nearest Neighbor-based Strategy (BNN) \cite{Laszlo2006} classifies or predicts groups of data points based on proximity. Correlation-driven Nonparametric Learning Approach (CORN) \cite{Lii2011} uses correlations to infer relationships between variables in a nonparametric learning context.

Our experiments use MIGT, EIIE, IMIT, FinRL, ES, TradeMaster, and SARL as DRL comparative frameworks. MIGT aims to maximize investment returns while ensuring the learning process's stability and reducing outlier impacts. EIIE has proven effective in managing cryptocurrency portfolios \cite{Jiang2017}, and we have transitioned this framework to the stock portfolio domain, optimizing it specifically for the stock market. IMIT models trading knowledge by emulating investor behaviour with logical descriptors and introduces the Rank-Invest model, which optimizes various evaluation metrics to preserve the diversity of these descriptors \cite{2018Investor}. FinRL offers a well-structured and robust framework for automated trading using DRL \cite{Liu2021}. ES integrates the best features of three actor-critic algorithms into a novel portfolio management framework \cite{Yang2020}. TradeMaster supports DRL-based quantitative trading (including portfolio management) and incorporates automated machine learning techniques to fine-tune hyperparameters for training reinforcement learning algorithms \cite{sun2023trademaster}. We will compare this with the PPO algorithm, TradeMaster\_PPO (TMP). Lastly, SARL enhances robustness to environmental uncertainties by using asset information and price trend predictions as additional states based on financial data \cite{Ye2020ReinforcementLearningBP}.

\begin{table}
        \centering
	\caption{Time composition of the training and test set}\label{tbl1}
	\begin{tabular}{l l l l}
		\toprule
		Dataset & Data purpose& Training data set& Back-test data interval  \\% Table header row
		\midrule
		1 & Back-test 2019 & 2016.01.01– 2018.12.31 & 2019.01.01– 2019.12.31  \\
		2 & Back-test 2020 & 2017.01.01– 2019.12.31 & 2020.01.01– 2020.12.31  \\
		3 & Back-test 2021 & 2018.01.01– 2020.12.31 & 2021.01.01– 2021.12.31  \\
  	4 & Back-test 2022 & 2019.01.01– 2021.12.31 & 2022.01.01– 2022.12.31  \\
   	5 & Back-test 2023 & 2020.01.01– 2022.12.31 & 2023.01.01– 2023.12.31  \\
		\bottomrule
	\end{tabular}
\end{table}

\begin{longtable}{l l l l l}
\caption{Results of the comparative experiments, where the best results for each metric are in bold.}\label{tbl2} \\

\toprule
Strategies& Cumulative returns & Sharpe ratio & Omega ratio & Sortino ratio \\  
\toprule
\multicolumn{5}{c}{Dataset 1}\\
\midrule
\pmb{MTS}  &\pmb{0.5203}  & \pmb{2.6872}  &  \pmb{1.5941} &  \pmb{4.0677}       \\
MIGT & 0.3844 & 1.7250 & 1.3399 & 2.7407 \\
EIIE & 0.1721 & 0.9518 & 1.1900 & 1.6544 \\
IMIT  & -0.0040 &	-0.0619 &	0.9895 &	-0.0826 \\
Finrl & 0.0911 & 0.4290 & 1.0768 & 0.8286 \\
ES & 0.1729 & 1.3562 & 1.2760 & 2.3045 \\
TMP & 0.2195 &	1.5178 &	1.3077 &	2.1441 \\
SARL & 0.2018 &	1.3662 &	1.2584 &	1.9301\\
BEST & -0.1045 & -0.5067 & 0.9148 & -0.5422 \\
BNN & -0.4275 & -3.0979 & 0.5911 & -3.5436 \\
CORN & -0.3728 & -2.3876 & 0.6487 & -2.6570 \\
CWMR & -0.2826 & -1.2873 & 0.7990 & -1.4729 \\
OLMAR & -0.3259 & -1.4806 & 0.7774 & -1.7227 \\
PAMR & -0.2873 & -1.3099 & 0.7957 & -1.4983 \\
RMR & -0.3087 & -1.3997 & 0.7864 & -1.6485 \\
TCO & -0.0397 & -0.3270 & 0.9465 & -0.2088 \\
WMAMR & -0.0570 & -0.2533 & 0.9586 & -0.1702 \\

\toprule
\multicolumn{5}{c}{Dataset 2}\\  % Table header row
\midrule
\pmb{MTS}     & \pmb{0.2793}      & \pmb{0.7920}      & \pmb{1.1561}     & \pmb{1.2282}       \\
MIGT & 0.2444 & 0.7069 & 1.1464 & 1.1809 \\
EIIE & 0.1469 & 0.4750 & 1.1046 & 0.7951 \\
IMIT & -0.2527 &	-0.5109 &	0.9070 &	-0.6719 \\
Finrl & 0.0194 & 0.1966 & 1.0405 & 0.3743 \\
ES & 0.0565 & 0.2247 & 1.0432 & 0.5049 \\
TMP & 0.0783 &	0.3042 &	1.0642 &	0.4260 \\
SARL & 0.0856 &	0.3246 &	1.0673 &	0.4656 \\
BEST & -0.4428 & -0.6463 & 0.8807 & -0.9021 \\
BNN & -0.3111 & -0.4830 & 0.9025 & -0.6816 \\
CORN & -0.0453 & 0.0067 & 1.0015 & 0.1142 \\
CWMR & -0.6093 & -1.3362 & 0.7460 & -1.6241 \\
OLMAR & -0.2917 & -0.2150 & 0.9550 & -0.2231 \\
PAMR & -0.6157 & -1.3546 & 0.7428 & -1.6435 \\
RMR & -0.2556 & -0.2047 & 0.9572 & -0.2027 \\
TCO & -0.5006 & -0.9751 & 0.7987 & -1.1868 \\
WMAMR & -0.1504 & 0.0469 & 1.0103 & 0.1264 \\

\toprule
\multicolumn{5}{c}{Dataset 3}\\  % Table header row
\midrule
\pmb{MTS}      & \pmb{0.2964}            & \pmb{1.4722}      & \pmb{1.2729}     & \pmb{2.2665}       \\
MIGT & 0.2834 & 1.3013 & 1.2413 & 2.1892 \\
EIIE & 0.1337 & 0.5482 & 1.1008 & 0.9786 \\
IMIT & 0.1496  &	0.7673 &	1.1344 &	1.1064 \\
Finrl & 0.1181 & 0.5668 & 1.0969 & 1.0844 \\
ES & 0.1608 & 0.8906 & 1.1600 & 1.6379 \\
TMP & 0.1690  &	1.1603 &	1.2127 &	1.6797 \\
SARL & 0.1472 &	0.9339 &	1.1674 &	1.3345 \\
BEST & -0.3790 & -1.5842 & 0.7493 & -1.9076 \\
BNN & 0.1245 & 0.4810 & 1.0831 & 0.9592 \\
CORN & -0.2696 & -1.1886 & 0.8091 & -1.3683 \\
CWMR & -0.4195 & -2.2341 & 0.6788 & -2.7035 \\
OLMAR & -0.2401 & -1.0747 & 0.8328 & -1.3266 \\
PAMR & -0.4162 & -2.2112 & 0.6814 & -2.6759 \\
RMR & -0.2913 & -1.3316 & 0.7973 & -1.6407 \\
TCO & -0.1341 & -0.9244 & 0.8529 & -1.0772 \\
WMAMR & -0.2190 & -1.0236 & 0.8383 & -1.2724 \\

\toprule
\multicolumn{5}{c}{Dataset 4}\\  % Table header row
\midrule
\pmb{MTS}      & \pmb{0.2642}            & \pmb{0.7117}      & \pmb{1.1480}     & \pmb{1.2091}       \\
MIGT & -0.2040 & -0.7208 & 0.8866 & -0.9893 \\
EIIE & -0.1057  & -0.4882  & 0.9229  & -0.6619 \\
IMIT & -0.0637  & -0.3576  & 0.9432  & -0.4890 \\
Finrl & -0.0545  & -0.3319  & 0.9462  & -0.4668 \\
ES & -0.1085  & -0.6590  & 0.8970  & -0.9089 \\
TMP & -0.1592  & -0.8342  & 0.8717  & -1.1347 \\
SARL & -0.0387  & -0.1086  & 0.9818  & -0.1646 \\
BEST & -0.2110  & -0.6759  & 0.8931  & -0.9675 \\
BNN & 0.1726  & 0.5517  & 1.0997  & 0.8297 \\
CORN & -0.2315  & -0.9673  & 0.8529  & -1.2907 \\
CWMR & -0.5668  & -2.2654  & 0.6800  & -2.8476 \\
OLMAR & -0.1069  & -0.1935  & 0.9680  & -0.2843 \\
PAMR & -0.5662  & -2.2655  & 0.6800  & -2.8475 \\
RMR & -0.3387  & -0.9484  & 0.8514  & -1.3098 \\
TCO & -0.2591  & -1.0252  & 0.8430  & -1.3720 \\
WMAMR & -0.3489  & -1.2011  & 0.8181  & -1.5930 \\

\toprule
\multicolumn{5}{c}{Dataset 5}\\  % Table header row
\midrule
\pmb{MTS}      & \pmb{0.2671}            & \pmb{0.9517}      & \pmb{1.1837}     & \pmb{1.4831}       \\
MIGT & 0.0304 & 0.0861 & 1.0142 & 0.1163 \\
EIIE & 0.1614  & 0.8296  & 1.1435  & 1.2180  \\
IMIT & 0.1406  & 0.9273  & 1.1675  & 1.3671  \\
Finrl & 0.1015  & 0.6565  & 1.1114  & 0.9628  \\
ES & 0.1129  & 0.7570  & 1.1301  & 1.1137  \\
TMP & 0.0898  & 0.5647  & 1.0950  & 0.7923  \\
SARL & 0.0413  & 0.1490  & 1.0255  & 0.2164  \\
BEST & -0.2295  & -1.0029  & 0.8461  & -1.4401  \\
BNN & 0.0653  & 0.2618  & 1.0472  & 0.4065  \\
CORN & -0.1253  & -0.5593  & 0.9089  & -0.7643  \\
CWMR & -0.3346  & -1.7581  & 0.7485  & -2.3242  \\
OLMAR & 0.0698  & 0.2746  & 1.0480  & 0.4445  \\
PAMR & -0.3339  & -1.7464  & 0.7496  & -2.3137  \\
RMR & 0.0350  & 0.1560  & 1.0269  & 0.2493  \\
TCO & -0.0628  & -0.4460  & 0.9303  & -0.6217  \\
WMAMR & -0.2068  & -0.8508  & 0.8636  & -1.2087  \\
\bottomrule

\end{longtable}

\begin{figure}
	\centering
	\includegraphics[width=12cm]{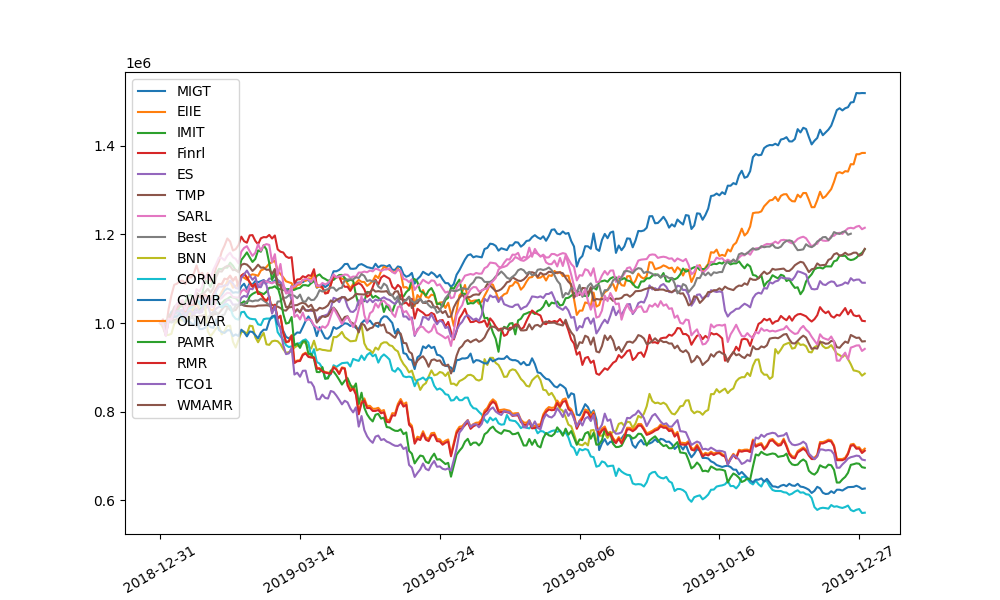}
	\caption{Results of the comparative Experiments (Dataset 1)}\label{fig2019}
\end{figure}

\begin{figure}
	\centering
	\includegraphics[width=12cm]{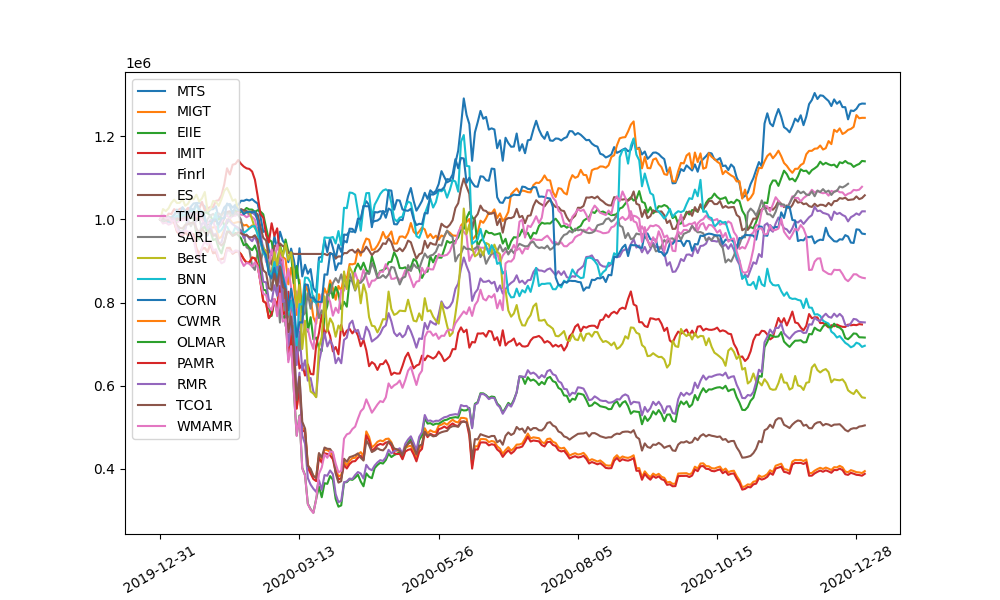}
	\caption{Results of the comparative Experiments (Dataset 2)}\label{fig2020}
\end{figure}

\begin{figure}
	\centering
	\includegraphics[width=12cm]{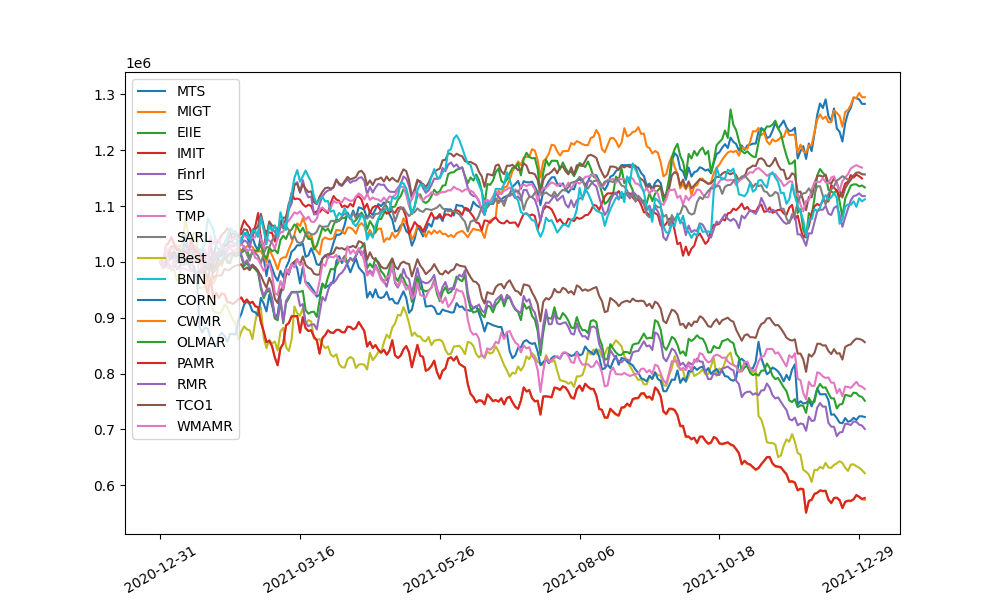}
	\caption{Results of the comparative Experiments (Dataset 3)}\label{fig2021}
\end{figure}

\begin{figure}
	\centering
	\includegraphics[width=12cm]{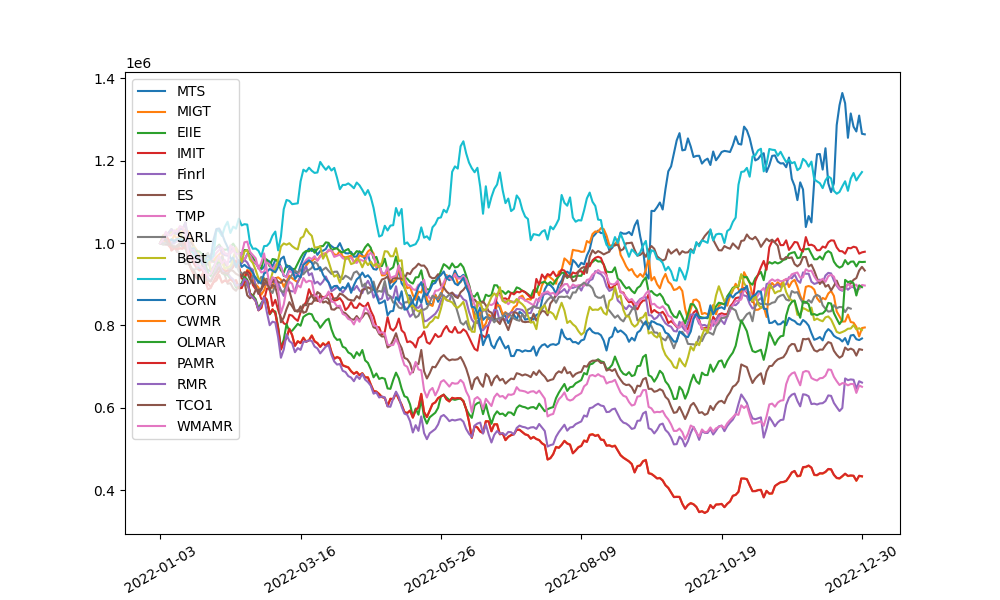}
	\caption{Results of the comparative Experiments (Dataset 4)}\label{fig2022}
\end{figure}

\begin{figure}
	\centering
	\includegraphics[width=12cm]{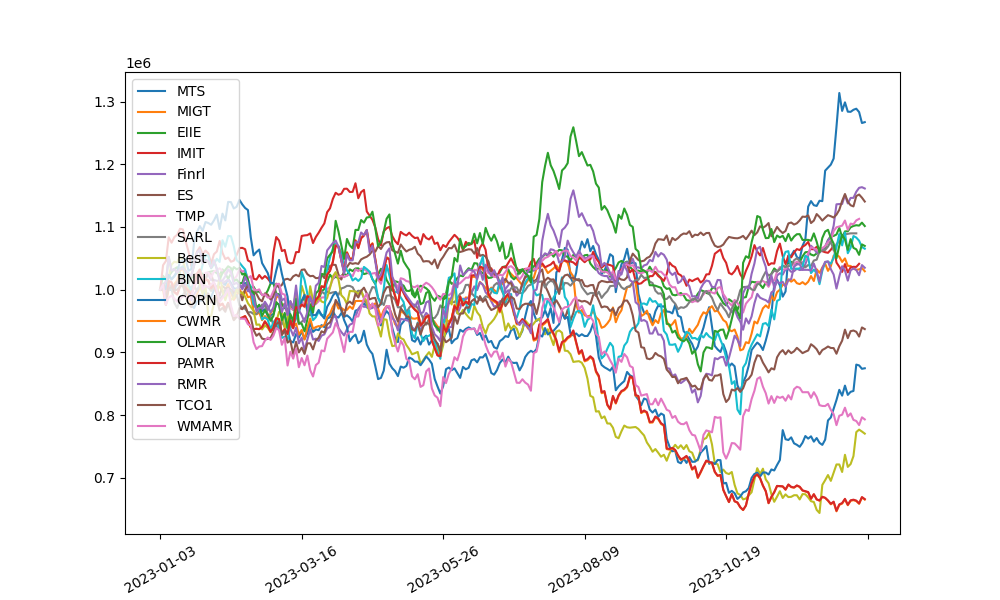}
	\caption{Results of the comparative Experiments (Dataset 5)}\label{fig2023}
\end{figure}

The experimental results in Table \ref {tbl2} and Figure \ref{fig2019} to \ref{fig2023} provide evidence for the efficacy of the MTS strategy. Across five diverse datasets, MTS consistently outperforms existing strategies, including traditional algorithms like OLMAR and PAMR, as well as state-of-the-art machine learning approaches such as MIGT and EIIE. If we average the results of the five experiments, our MTS achieves a cumulative return of 0.3255, which is 30.67$\%$ higher than the average of the second-best performance in each experiment, which is 0.2492. The consistent outperformance of MTS across all datasets and metrics underscores the synergy among its components. ICVaR provides a risk-managed foundation, allowing aggressive pursuit of returns within defined risk tolerances. The time-aware attention mechanism identifies temporal opportunities within this risk framework, enhancing return generation. Finally, the parallel DRL executes the strategy robustly, adapting to market trends that may themselves be more accurately identified due to time-awareness.

This combined effect is especially noticeable in Dataset 1, where MTS shows the most significant performance across all metrics. One possible reason is that this dataset represents a market environment characterized by high volatility (benefiting ICVaR), strong temporal effects (leveraged by time-aware attention), or distinct trends (captured by parallel DRL). Conversely, in Dataset 2, where the performance in Omega and Sortino ratios are smallest (despite still positive at 0.84$\%$ and 4.01$\%$ respectively), the market likely presents fewer exploitable inefficiencies or less pronounced trends.

\begin{figure}
	\centering
	\includegraphics[width=12cm]{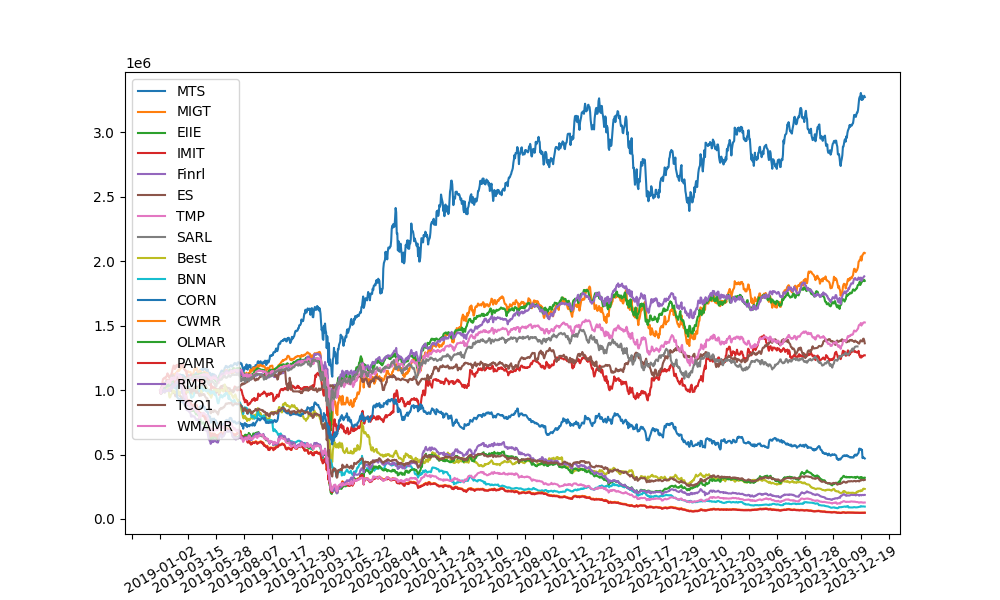}
	\caption{Results of the comparative Experiments (Dataset 1-5)}\label{fig201923}
\end{figure}

To validate MTS's performance over the long term, we use the 2019-2023 five-year data, i.e., combining datasets 1 to 5 (Table \ref{tbla} and Figure \ref{fig201923}), as a comparative experiment for the test set. The experimental results demonstrate that MTS significantly outperforms other strategies across all evaluated metrics over the five-year period. Specifically, MTS shows a cumulative return of 2.2739, which is 115$\%$ higher than the second-best strategy, MIGT, with a cumulative return of 1.0639. Additionally, MTS boasts the highest Sharpe, Omega, and Sortino ratios, indicating its superior risk-adjusted performance and stability in varying market conditions. This suggests that MTS is a robust strategy for long-term investment, effectively maximizing returns while minimizing downside risks.

\begin{table}
	\caption{Five years of data from 2019-2023 were used as a test set to compare the results of the experiments, with the best results for each metric given in bold.}\label{tbla}
	\centering
	\begin{tabular}{l l l l l l}
		\toprule
		Strategies&  Cumulative returns & Sharpe ratio & Omega ratio & Sortino ratio \\  
		\midrule
        \pmb{MTS}  & \pmb{2.2739} & \pmb{0.9149} & \pmb{1.1875} & \pmb{1.3319} \\
        MIGT & 1.0639 & 0.5505 & 1.1166 & 0.8024 \\
        EIIE & 0.8832 & 0.6816 & 1.1302 & 0.9511 \\
        IMIT & 0.3634 & 0.2773 & 1.0507 & 0.3901 \\
        Finrl & 0.8476 & 0.5114 & 1.1205 & 0.7453 \\
        ES & 0.5251 & 0.3761 & 1.0786 & 0.5268 \\
        TMP & -0.9866 & -0.3040 & 0.8750 & -0.3181 \\
        SARL & 0.2703 & 0.2077 & 1.0422 & 0.3032 \\
        BEST & -0.7668 & -0.6925 & 0.8762 & -0.9851 \\
        BNN & -0.9038 & -1.2821 & 0.7823 & -1.6851 \\
        CORN & -0.5303 & -0.4096 & 0.9262 & -0.5770 \\
        CWMR & -0.9500 & -1.5147 & 0.7383 & -1.9473 \\
        OLMAR & -0.6797 & -0.4311 & 0.9194 & -0.5848 \\
        PAMR & -0.9522 & -1.5347 & 0.7351 & -1.9701 \\
        RMR & -0.8131 & -0.7236 & 0.8691 & -0.9587 \\
        TCO & -0.6960 & -0.6799 & 0.8646 & -0.9056 \\
        WMAMR & -0.8721 & -0.9859 & 0.8237 & -1.2740 \\

		\bottomrule
	\end{tabular}
\end{table}

\subsection{Ablation Study and Results Analysis}\label{}
\subsubsection{Overall Ablation Study}\label{}
To evaluate the individual contributions of each component in our MTS strategy, we conducted an ablation study across five datasets representing different market conditions from 2019 to 2023. We compared the full MTS model against three variants, each with one key component removed:

\begin{itemize}
    \item MTS\_w/o\_ICVaR: MTS without the ICVaR at Risk component
    \item MTS\_w/o\_Time-Awareness: MTS without the time-aware attention mechanism
    \item MTS\_w/o\_Short-Selling: MTS without the ability to short-selling
\end{itemize}

\begin{longtable}{l l l l l}
\caption{Results of ablation study, where the best results for each metric are in bold.}\label{tbab} \\

\toprule
Strategies& Cumulative returns & Sharpe ratio & Omega ratio & Sortino ratio \\  
\toprule
\multicolumn{5}{c}{Dataset 1}\\
\midrule
\pmb{MTS}  &\pmb{0.5203}  & \pmb{2.6872}  &  \pmb{1.5941} &  \pmb{4.0677}       \\
MTS\_w/o\_ICVaR & 0.4697 & 2.4197 & 1.5146 & 3.6508  \\
MTS\_w/o\_Time-Aware  & 0.2763 & 1.4617 & 1.2767 & 2.0362  \\
MTS\_w/o\_Short-Selling & 0.4280 & 2.2459 & 1.4772 & 3.3669  \\

\toprule
\multicolumn{5}{c}{Dataset 2}\\  % Table header row
\midrule
\pmb{MTS}     & \pmb{0.2793}      & \pmb{0.7920}      & \pmb{1.1561}     & \pmb{1.2282}       \\
MTS\_w/o\_ICVaR & 0.2529 & 0.7426 & 1.1397 & 1.1250  \\
MTS\_w/o\_Time-Aware  & 0.2444 & 0.6892 & 1.1392 & 0.9588  \\
MTS\_w/o\_Short-Selling & 0.2505 & 0.7585 & 1.1517 & 1.1075  \\

\toprule
\multicolumn{5}{c}{Dataset 3}\\  % Table header row
\midrule
MTS  & 0.2964           & 1.4722    & 1.2729    & 2.2665      \\
MTS\_w/o\_ICVaR & 0.2848 & 1.4056 & 1.2609 & 2.1674  \\
MTS\_w/o\_Time-Awareness  & 0.1209 & 0.4674 & 1.0873 & 0.6585  \\
\pmb{MTS\_w/o\_Short-Selling} & \pmb{0.3097} & \pmb{1.5806} & \pmb{1.3009} & \pmb{2.3649}  \\

\toprule
\multicolumn{5}{c}{Dataset 4}\\  % Table header row
\midrule
\pmb{MTS}      & \pmb{0.2642}            & \pmb{0.7117}      & \pmb{1.1480}     & \pmb{1.2091}       \\
MTS\_w/o\_ICVaR & 0.1862 & 0.5940 & 1.1130 & 0.9461  \\
MTS\_w/o\_Time-Awareness  & 0.0477 & 0.2260 & 1.0392 & 0.3386  \\
MTS\_w/o\_Short-Selling & -0.1662 & -0.8415 & 0.8715 & -1.1548  \\

\toprule
\multicolumn{5}{c}{Dataset 5}\\  % Table header row
\midrule
\pmb{MTS}      & \pmb{0.2671}            & \pmb{0.9517}      & \pmb{1.1837}     & \pmb{1.4831}       \\
MTS\_w/o\_ICVaR & 0.2319 & 0.9012 & 1.1553 & 1.3294  \\
MTS\_w/o\_Time-Awareness  & 0.2528 & 1.3171 & 1.2396 & 1.9980  \\
MTS\_w/o\_Short-Selling & 0.1771 & 0.9803 & 1.1745 & 1.3857  \\

\bottomrule

\end{longtable}

The ablation study reveals that the time-aware attention mechanism has the most significant impact on the performance metrics across the datasets, with its omission leading to the largest decreases. On average, the metrics decreased by about 43.7$\%$ for cumulative returns, 50.4$\%$ for the Sharpe ratio, 6.5$\%$ for the Omega ratio, and 54.8$\%$ for the Sortino ratio when the time-aware component was removed. In contrast, the ICVaR component had a smaller impact, with average decreases of 7.2$\%$, 4.6$\%$, 1.6$\%$, and 8.7$\%$ in the respective metrics. However, in Dataset 4, where the market exhibits a continuous downward trend, the ability to short-sell demonstrated its importance more prominently. Removing the short-selling capability resulted in more significant decreases of 81.2$\%$ for cumulative returns, 188.1$\%$ for the Sharpe ratio, 12.9$\%$ for the Omega ratio, and 273.2$\%$ for the Sortino ratio. This highlights the critical role of short-selling in managing and mitigating losses during market downturns.

\subsubsection{Risk Management through ICVaR}\label{}
At the core of MTS's success is its novel risk management framework, ICVaR, which incorporates adjustable risk preferences into the reward function. The impact of this contribution is most evident in the risk-adjusted performance metrics. Across all datasets from Table \ref {tbl2}, MTS achieves higher Sharpe ratios, outperforming the next-best strategy by a margin ranging from 12.04$\%$ (Dataset 2) to an impressive 55.78$\%$ (Dataset 1). This consistent improvement in risk-adjusted returns suggests that ICVaR effectively balances the pursuit of high returns with the imperative of risk mitigation.

Furthermore, the Sortino ratios, which focus on downside deviations, demonstrate a greater advantage. MTS's Sortino ratios are 4.01$\%$ to 48.42$\%$ higher than the second-best strategy, with the peak difference again in Dataset 1. This underscores ICVaR's particular strength in mitigating downside risk, which is a crucial consideration in portfolio management. The fact that these improvements are most pronounced in Dataset 1 suggests that ICVaR's benefits are amplified in high-risk or volatile market environments, allowing MTS to capture upside potential without exposing the portfolio to undue risk.

In the overall ablation study, the ICVaR component consistently improves performance across all datasets. Its impact is particularly notable in Dataset 4, where removing ICVaR significantly drops all metrics (e.g., Sharpe ratio decreases from 0.71168 to 0.59399). This suggests that ICVaR plays a crucial role in risk management, especially in challenging market conditions.

\subsubsection{Capturing Temporal Market Inefficiencies}\label{}
MTS has a redesigned encoder and attention mechanism, incorporating temporal market characteristics such as the weekend and turn-of-the-month effects. This time-aware attention mechanism translates into consistent outperformance in cumulative returns, a primary measure of strategy effectiveness. MTS surpasses the second-best strategy by 4.58$\%$ to 35.33$\%$, a substantial edge in the competitive field of algorithmic trading.

The degree of outperformance in cumulative returns varies across datasets, from 4.58$\%$ in Dataset 3 to 35.33$\%$ in Dataset 1. This variability suggests that the importance of temporal features differs across market environments. MTS's time-aware design provides a significant advantage in markets with more pronounced effects (Dataset 1). Conversely, the advantage is more modest yet still present in markets where these effects are less prominent (Dataset 3). This adaptability is a key advantage of MTS, enabling it to extract temporally correlated features. It shows varying impacts across datasets in the overall ablation study. Its contribution is most significant in Datasets 1 and 4, where its removal leads to substantial decreases in all metrics. For instance, in Dataset 1, the Sharpe ratio drops from 2.68723 to 1.46166 without this component. This indicates that the time-aware mechanism is particularly effective in capturing temporal market inefficiencies in certain market conditions.

\begin{figure}
	\centering
	\includegraphics[width=10cm]{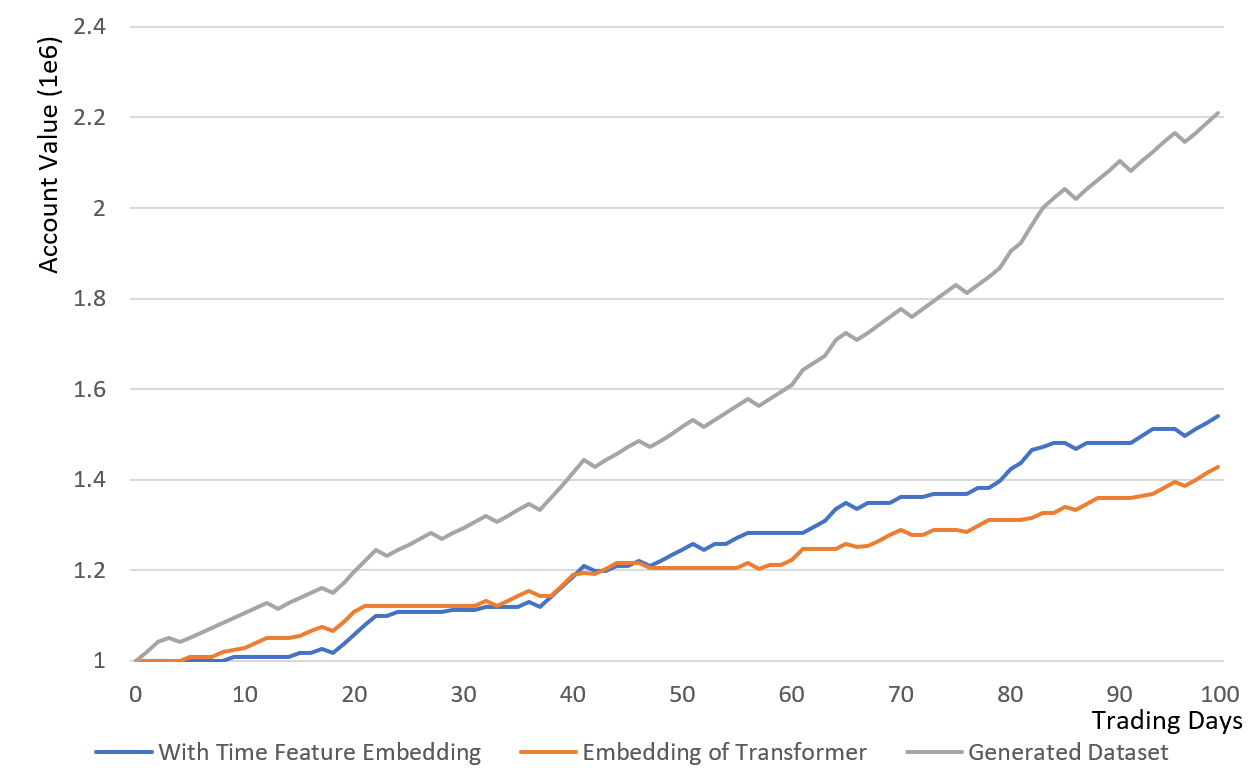}
	\caption{Comparison of markets with strong weekend effect and turn-of-the-month effect}\label{figcu}
\end{figure}

To verify the role of time feature embedding on the returns of portfolio strategies, we generate a set of data (in Figure \ref{figcu}) with a strong weekend effect and turn-of-the-month effect, that is, add random fluctuations on the weekend and early month of a simulated smooth market for testing. From Figure \ref{figcu}, we find that starting from the 47th trading day, the DRL strategy with time feature embedding shows an advantage, indicating that it more fully extracts the time feature in the stock market.

\subsubsection{Adaptive Transaction through Parallel DRL}\label{}
MTS's parallel DRL for market condition determination complements the previous two by adapting trading strategies based on detected market trends. This component's impact is most evident in challenging market conditions, exemplified by Dataset 4. Some strategies that achieved positive returns in other datasets suffered losses in Dataset 4. In contrast, our MTS strategy maintained impressive returns.

This performance in adverse conditions can be attributed to the parallel DRL's adaptive trading rules, such as limiting buying during downtrends and managing short-selling. These rules, informed by real-time trend analysis, contribute significantly to downside protection, as reflected in the superior Sortino ratios. Moreover, we hypothesize that the trend identification itself benefits from the time-aware attention mechanism, creating a synergistic effect that enhances MTS's adaptability.

To test the profitability of our strategy that includes short-selling in a declining market, we have simulated an extreme market that declines by 5$\%$ per trading day. In the experiment, the short-selling cap $u$ is set to 1e5, 1e4, and 0, respectively.

\begin{figure}
	\centering
	\includegraphics[width=10cm]{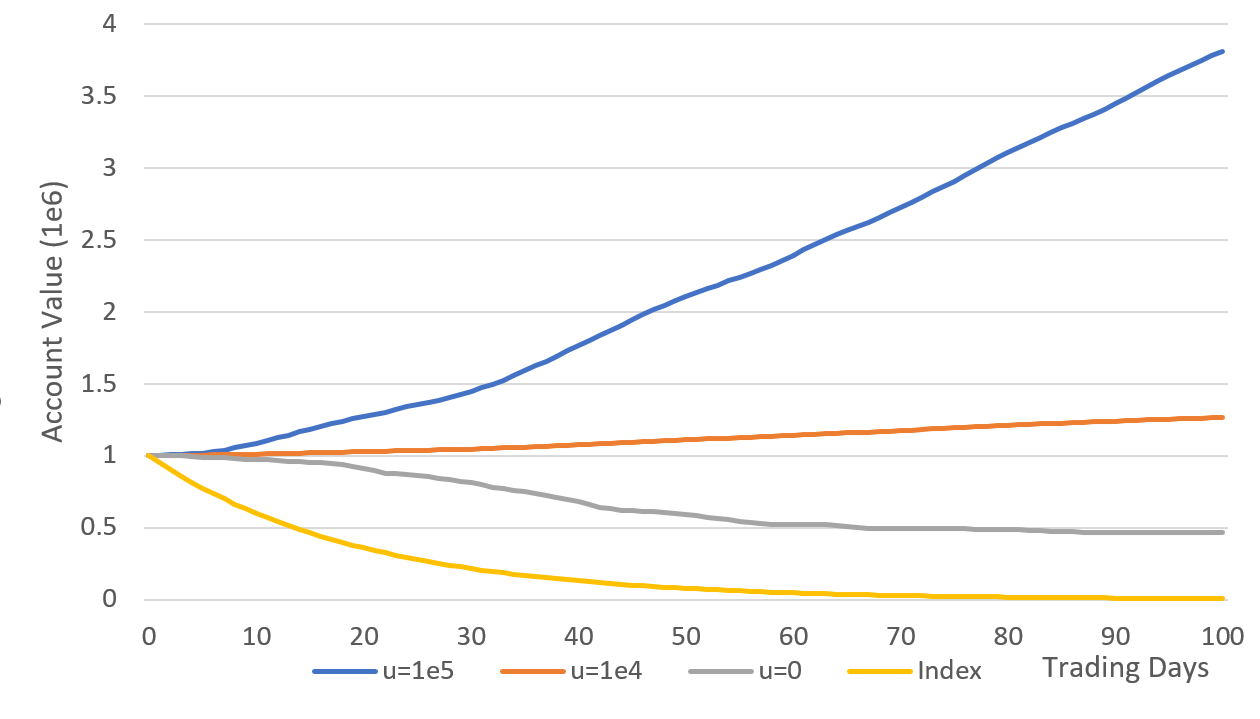}
	\caption{Comparison of market declining by 5$\%$ per day}\label{figc}
\end{figure}

The experimental results show that when $u$ is 1e5, the portfolio strategy obtains a substantial return of 280$\%$  using the short-selling strategy, and this return also reaches 26.58$\%$  when $u$ is 1e4. When without short-selling, the long investment trade loses about half its value in this extreme market, but it is still a stopgap compared to the simulated market index's 99.41 $\%$  loss. In Dataset 4 of the overall ablation study (Table \ref {tbab}), which likely represents a bear market, removing short-selling capability leads to negative returns (-0.16623 vs. 0.26417 with short-selling).

We also compared the impact of applying short-selling in a simulated market of up to 5$\%$ per day. Although the returns are lower than those of the long strategy (approximately 5.41$\%$), no declines have occurred. This suggests that the strategy that allowed short-selling does not carry out observable short-selling in this consistently rising market, and the short-selling amount limit $u$ controlled by the virtual sub-strategy remains at zero or very low levels. The lower returns from applying the short-selling strategy here are mainly due to our accompanying module on limiting some of the buying to reduce risk. Similar results appear in Dataset 4 of the overall ablation study, where the strategy performs slightly better without short-selling.

\section{Conclusions and Future Work}\label{s5}
This study proposes a novel strategy, MTS, for portfolio management using deep reinforcement learning. Our approach addresses key challenges in algorithmic trading through three innovations: ICVaR for dynamic risk management, a time-aware attention mechanism to capture temporal market inefficiencies, and a parallel DRL framework for adaptive trading with short-selling control. The experimental results, conducted over five diverse datasets spanning from 2019 to 2023, demonstrate the superior performance of MTS compared to a wide range of existing strategies, including both traditional algorithms and state-of-the-art machine learning approaches. MTS consistently outperformed other strategies across multiple performance metrics, including cumulative returns, Sharpe ratio, Omega ratio, and Sortino ratio.

Despite its promising performance, our MTS strategy has three major limitations. First, our current implementation primarily focuses on equity markets and, like existing strategies, lacks comprehensive support for diverse financial markets. This limitation restricts the strategy's applicability across the broader financial landscape. Second, while the reinforcement learning method we use PPO is effective, it is not specifically tailored for portfolio management tasks, potentially missing out on domain-specific optimizations. Third, although we incorporate risk management through ICVaR, our model does not fully utilize the wide array of advanced mathematical models available in financial theory, which could potentially enhance the strategy's risk assessment and decision-making capabilities.

We propose three directions for future work to address these limitations. First, we aim to extend MTS to support a broader range of financial markets simultaneously, including bonds and foreign exchange. This expansion would involve adapting our model to handle these markets' unique characteristics and data structures, potentially leading to a more versatile and robust strategy. Second, we plan to develop a novel reinforcement learning algorithm for portfolio management tasks instead of existing algorithms like PPO, DDPG and SAC. This new algorithm could incorporate financial domain knowledge directly into its architecture, potentially including custom policy and value networks that better capture the complexities of financial markets.  Third, we intend to incorporate more advanced mathematical models from financial theory into our framework. This could include integrating stochastic volatility models for improved risk estimation or incorporating regime-switching models to capture market dynamics better. These enhancements would improve our approach's theoretical foundation and potentially improve real-world performance across various market conditions.

% \end{thebibliography}
\bibliographystyle{SageV}
\bibliography{sn-bibliography}
\end{document}